\documentclass[letterpaper, 10 pt, conference]{ieeeconf}  

\IEEEoverridecommandlockouts                              

\usepackage{graphicx, tipa}
\usepackage{amsmath,amssymb,amsfonts}
\usepackage[caption=false]{subfig}
\usepackage[ruled,linesnumbered, noend]{algorithm2e}
\usepackage{dirtytalk}
\usepackage{hyperref}
\usepackage{gensymb}
\usepackage{tabularx}
\usepackage{multirow}
\usepackage[short]{optidef}
\usepackage{booktabs}
\usepackage{siunitx}
\usepackage{titlesec}
\usepackage{makecell}
\newcounter{cases}
\newcounter{subcases}[cases]

\makeatletter
\newcommand{\removelatexerror}{\let\@latex@error\@gobble}
\makeatother
\SetKwComment{Comment}{$\triangleright$\ }{}

\newcommand\Tstrut{\rule{0pt}{2.0ex}}         

\title{\bf \huge 
Heuristic-based Incremental Probabilistic Roadmap for Efficient UAV Exploration in Dynamic Environments
}

\author{Zhefan Xu\footnotemark*, Christopher Suzuki\footnotemark*, Xiaoyang Zhan, and Kenji Shimada 
\thanks{*The authors contributed equally.
    \newline{\indent Zhefan Xu, Christopher Suzuki, Xiaoyang Zhan, and Kenji Shimada are with the Department of Mechanical Engineering, Carnegie Mellon University, 5000 Forbes Ave, Pittsburgh, PA, 15213, USA.}
        {\tt\small zhefanx@andrew.cmu.edu}}%
}

\begin{document}

\maketitle
\thispagestyle{empty}
\pagestyle{empty}

\begin{abstract}
Autonomous exploration in dynamic environments necessitates a planner that can proactively respond to changes and make efficient and safe decisions for robots. Although plenty of sampling-based works have shown success in exploring static environments, their inherent sampling randomness and limited utilization of previous samples often result in sub-optimal exploration efficiency. Additionally, most of these methods struggle with efficient replanning and collision avoidance in dynamic settings. To overcome these limitations, we propose the Heuristic-based Incremental Probabilistic Roadmap Exploration (HIRE) planner for UAVs exploring dynamic environments. The proposed planner adopts an incremental sampling strategy based on the probabilistic roadmap constructed by heuristic sampling toward the unexplored region next to the free space, defined as the heuristic frontier regions. The heuristic frontier regions are detected by applying a lightweight vision-based method to the different levels of the occupancy map. Moreover, our dynamic module ensures that the planner dynamically updates roadmap information based on the environment changes and avoids dynamic obstacles. Simulation and physical experiments prove that our planner can efficiently and safely explore dynamic environments.
\end{abstract}

\section{Introduction}
Over the past decade, the integration of autonomous unmanned aerial vehicles (UAVs) into various industries for the rapid and reliable acquisition of information has become imperative. In scenarios where human workers coexist with UAVs in dynamic and unknown environments, the need for an online informative path-planning algorithm becomes evident. Such an algorithm is crucial for efficiently exploring unknown areas while maintaining the safety of human workers. Consequently, developing an efficient and safe exploration path planning algorithm is indispensable for overcoming the challenges posed by dynamic environments.

The online exploration challenge involves identifying a sequence of informative sensor positions \cite{first_nbv}. Early exploration strategies, pioneered with ground-based 2D robots, successfully gained insights into unknown regions by focusing on border regions (the frontiers) \cite{first_frontier}. While subsequent efforts extended these concepts to aerial robots \cite{2_stage}\cite{lu2020optimal}\cite{FUEL}, they can encounter computational bottlenecks in high-dimensional planning as the environment size grows. In contrast, sampling-based methods have gained favor in UAV exploration due to their computational efficiency and diverse information gain formulations \cite{NBVP}\cite{history-aware}\cite{AEP}\cite{incremental-rrt}. Typically, these methods employ cost-utility functions to assess exploration potentials at sampling nodes and compute paths using single-query planners like the Rapidly Exploring Random Tree (RRT) \cite{rrt} or its variations. However, the limitation of random sampling in a single iteration can result in incomplete coverage of mapped space nodes, hampering optimal viewpoint selection. Additionally, redundant computation may occur when evaluating previously sampled regions in subsequent iterations. Exploration being an iterative process, a multi-query planner like the Probabilistic Roadmap \cite{prm} is better suited with an efficient incremental sampling method to reduce computational costs and select viewpoints.

\begin{figure}
    \vspace{0.25cm}
    \centering
    \includegraphics[scale=0.395]{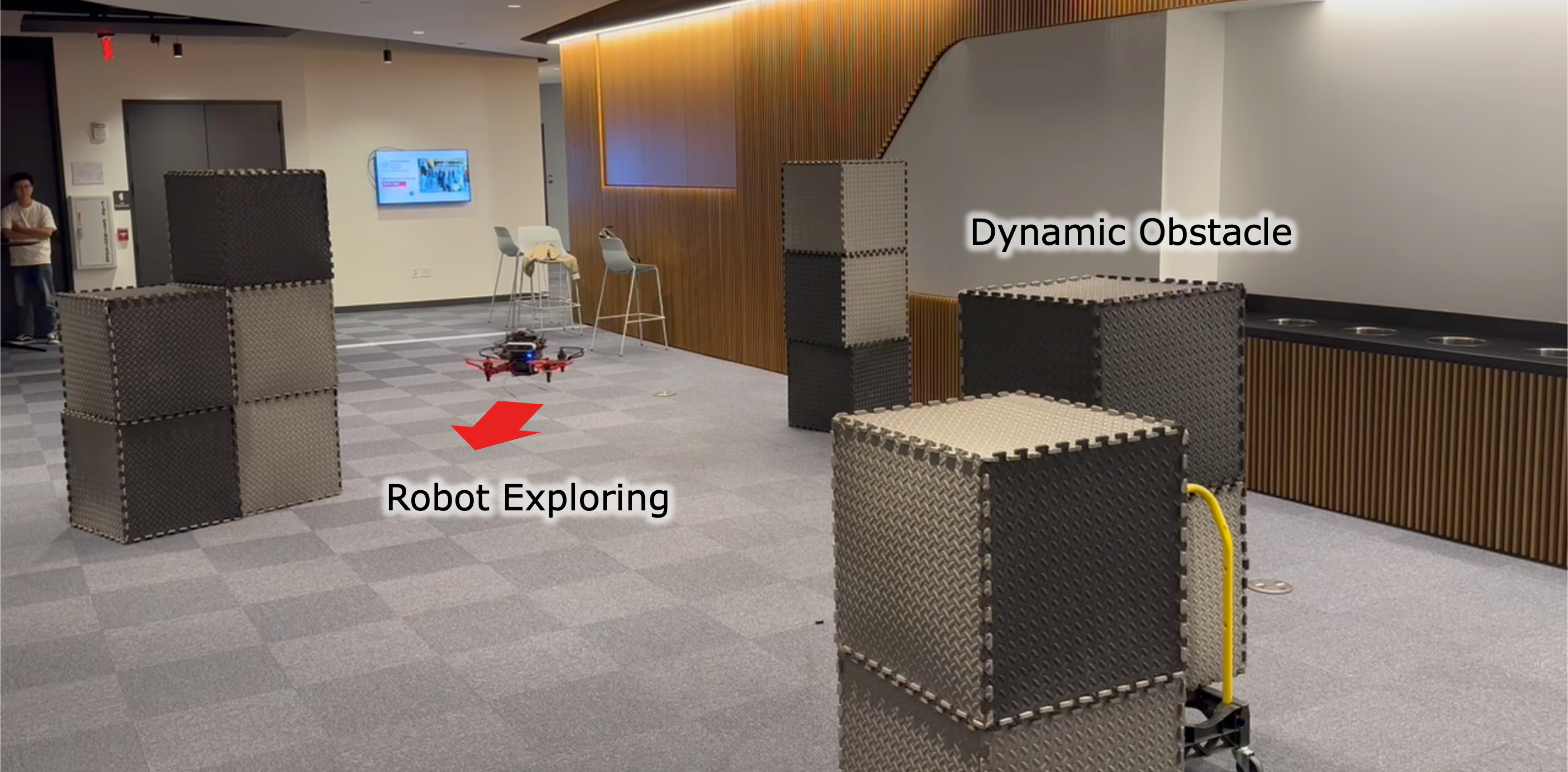}
    \caption{The UAV exploring the unknown environment with the proposed exploration planner. In this dynamic environment, the robot is conducting unknown space exploration while actively avoiding dynamic obstacles.}
    \label{intro figure}
\end{figure}

To address these issues, we propose a novel Heuristic-based Incremental Probabilistic Roadmap Exploration (HIRE) planner for exploring dynamic environments. The proposed planner employs an incremental sampling strategy, utilizing a probabilistic roadmap constructed through heuristic sampling near unexplored regions adjacent to free space, defined as heuristic frontier regions. These frontier regions are identified using a lightweight vision-based technique applied to various levels of the occupancy map. Furthermore, our dynamic module ensures that the planner dynamically updates the roadmap information in response to environmental changes and effectively avoids dynamic obstacles. Simulation and physical experiments conclusively demonstrate that our planner can efficiently and safely explore dynamic environments. The main contributions of this work are:
\begin{itemize}
    \item \textbf{Heuristic Incremental Roadmap Construction:}  The proposed algorithm applies a lightweight vision-based method to identify heuristic frontier regions across various levels of the map and adopts our heuristic sampling strategy for incremental roadmap construction.
    \item \textbf{Dynamic Module for Node Update and Collision Avoidance:}  Our dynamic module guarantees node information updates in dynamic environments while ensuring safe collision avoidance for dynamic obstacles.
    \item \textbf{High Exploration Efficiency:} We conducted comprehensive comparison experiments against state-of-the-art benchmark algorithms. The results demonstrate that our proposed planner is able to achieve higher exploration efficiency in different simulation environments.
\end{itemize}

\section{Related Work}
There are two major categories of autonomous exploration algorithms: the frontier-based method and the sampling-based method. Early frontier exploration has demonstrated high efficiency in 2D ground robots \cite{first_frontier}. The idea of using frontiers to guide exploration is later extended into UAVs with fast speed \cite{rapid_frontier} by selecting the goal frontier from its sensor field of view and minimizing the changes in its velocity. Meng et al.\cite{2_stage} adopt the idea of the frontier sampler for guiding the robot exploration. Lu et al. \cite{lu2020optimal} apply a clustering method to filter the frontiers and pick the optimal frontier based on the information gain. In \cite{FUEL}, the frontier information structure (FIS) is proposed, and their method utilizes three-step planning to achieve high exploration efficiency. While these works have demonstrated the success of frontier-based methods, it is important to note that the computational requirements for obtaining frontiers can grow exponentially with increasing environment size.

The sampling-based method has become more popular in recent years for UAV exploration. The Receding Horizon Next-Best-View (RH-NBV) planner provides a dependable solution for robot exploration \cite{NBVP}. It operates by expanding a tree from the robot's current position, selecting the best branch with the highest information gain, and having the robot execute the initial branch segment. Following the receding horizon manner, Papachristos et al. \cite{RHEM} minimize the localization and mapping uncertainties during the path selection. To enhance sampling efficiency and prevent the planner from becoming trapped in local minima, \cite{history-aware} stores a historical graph derived from previous samples to evaluate exploration potentials. In \cite{dang2018visual}, a two-stage planner is employed to optimize saliency gain, taking into account the visual saliency of various objects in the environment. In \cite{AEP}, the RH-NBV is combined with the frontier-based algorithm to prevent early termination in local minima within their autonomous exploration planner (AEP). The frontiers are cached to determine the exploration goals, and these cached nodes aid in estimating the information gain using Gaussian processes. To efficiently reuse sampling information, a graph-based planner \cite{GBP1} was initially proposed and later extended to support team robot exploration, taking into account wireless communication \cite{GBP2}. With a similar incremental strategy in \cite{OIPP}, their RRT*-based planner continuously expands and manages the tree with rewiring for path refinement, ensuring that it does not discard the other nodes not part of the best branch. Similarly, the incremental probabilistic roadmap is employed in \cite{DEP} to steer the exploration process. A group of methods \cite{GLocal}\cite{TARE}\cite{DSVP}\cite{UFOExplorer} adopts a combined approach, maintaining both global and local planning horizons to generate exploration trajectories. Additionally, recent works have incorporated semantic information \cite{semantic_unceratinty_exp}\cite{swap} and map prediction \cite{predcon} for enhancing exploration strategies.

\section{Problem Definition}
A bounded environment, $V_\text{env}\in\mathbb{R}^3$, consists of free space $V_\text{free}$ and occupied space $V_{\text{occ}}$. Without any occupancy information, the robot is required to explore the environment completely and build the voxel map $\mathcal{M}$ using its onboard sensor. At the beginning stage of the exploration, the entire environment is unknown except the robot's nearby region, the initial mapped space $\mathcal{M}_\text{init}$. With this initial map $\mathcal{M}_\text{init}$, the robot needs to iteratively generate collision-free trajectories to collect occupancy information using the onboard sensor. This process ends until the entire environment is completely explored as the environment map $\mathcal{M}_\text{env}$. Note that the final environment map should include all the free and occupied voxels except for some unreachable voxel (i.e., $\mathcal{M}_\text{env}=(V_\text{free} \cup V_\text{occ}) \setminus V_\text{ur}$). Due to the dynamic settings of the environment, there exist dynamic obstacles $O_\text{d}$. Some obstacles, such as moving tables, change locations infrequently, while some change positions quickly, such as pedestrians. As a result, the robot needs to update information based on these dynamic changes in the environment and avoid obstacles safely.

\section{Methodology} 
\begin{figure}
    \centering
    \includegraphics[scale=0.28]{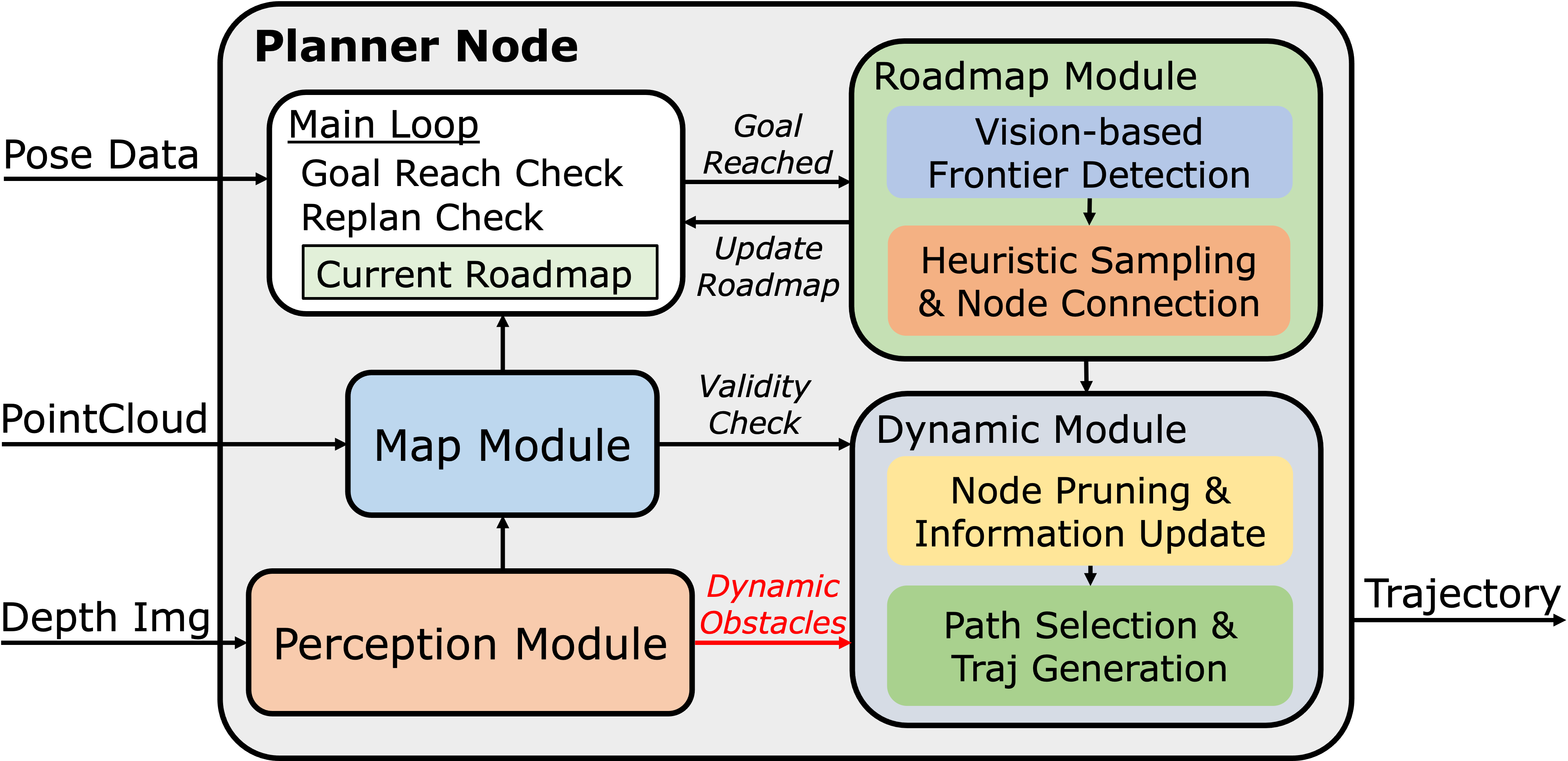}
    \caption{Planner overview. The system input comprises the pose, pointcloud, and depth image data. At the beginning of each planning iteration, the roadmap module incrementally constructs the roadmap. Consequently, the dynamic module will update node information and generate trajectories for exploration and collision avoidance based on map and perception data.}
    \label{system overview figure}
\end{figure}
This section introduces the proposed planner for UAVs exploring dynamic environments. There are four main steps in our planner within two modules: the roadmap module and the dynamic module, as shown in Fig. \ref{system overview figure}. The system input comprises the pose, pointcloud, and depth image data. The main loop performs the goal reach check and replan check based on map and dynamic obstacle information from the map and perception modules. After the main loop initializes a new planning iteration, the roadmap module first applies a vision-based detector to find the heuristic frontier regions (Sec. \ref{frontier detection section}) and then incrementally construct the probabilistic roadmap (Sec. \ref{roadmap construction section}). Within the dynamic module, the roadmap updates node information gain and prune invalid nodes based on the map validity check (Sec. \ref{information section}). Finally, the best view path is selected to generate collision-free trajectories for safe exploration (Sec. \ref{path section}).

\subsection{Vision-based Frontier Detection} \label{frontier detection section}
\begin{figure}
    \centering
    \includegraphics[scale=0.43]{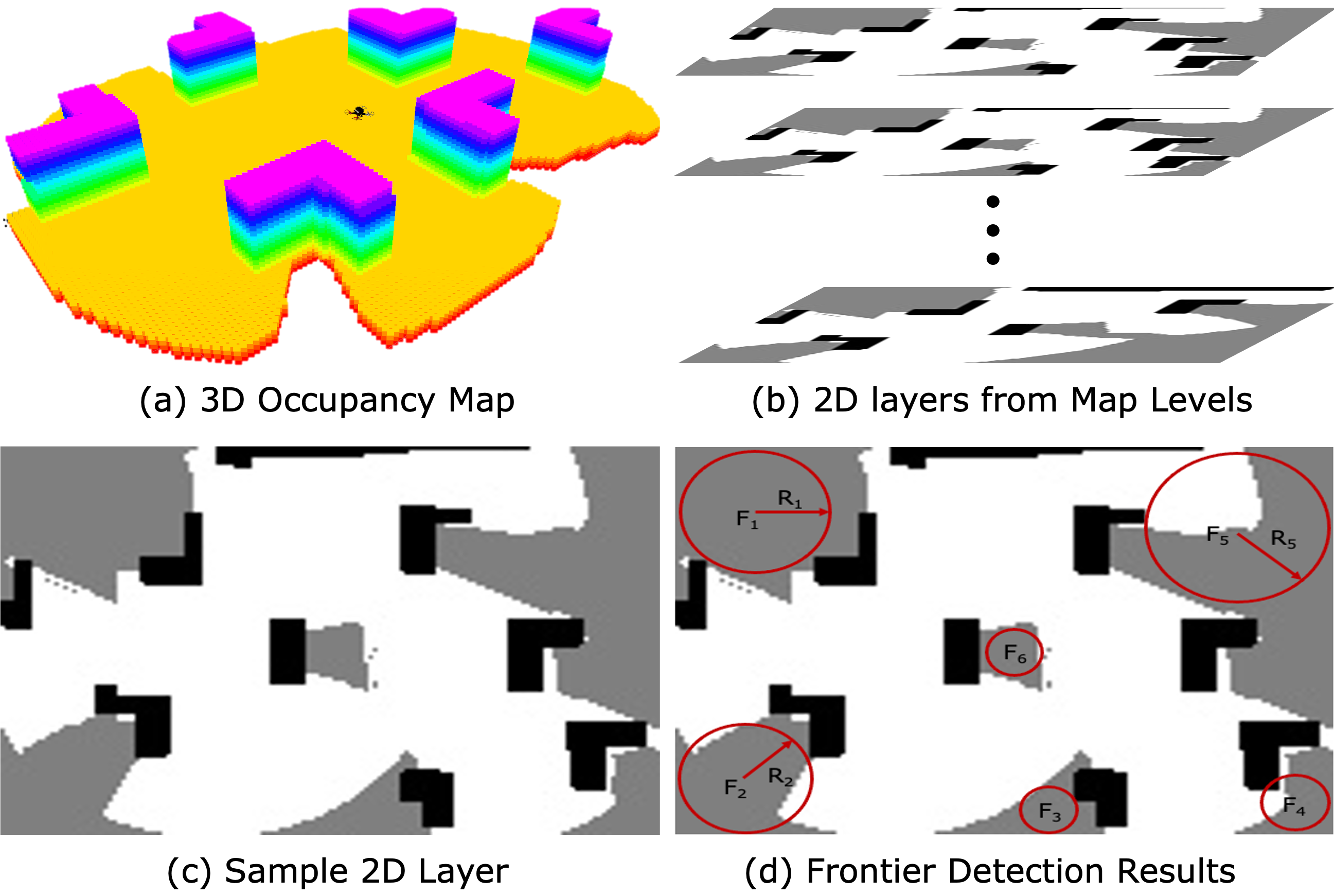}
    \caption{Illustration of the vision-based heuristic frontier detection method. (a) The 3D occupancy voxel map. (b) The 2D map from different levels of the map. (c) One sample 2D map converted to an image. (d) The frontier detection results by applying a vision-based detector to the 2D map image.}
    \label{frontier detection figure}
\end{figure}

\textbf{Heuristic Frontier Region:} To avoid complete randomness of node sampling and improve sampling efficiency, we utilize the heuristic frontiers for guiding sampling directions. The classic frontiers are defined as the boundaries separating free and unknown areas. Finding the classic frontiers involves traversing the entire 3D voxel map, which demands high computation. Differently, we define our heuristic frontier $\mathcal{F}$ as a circle with center $F$ and radius $R$ at the level $h$ where the circle should mainly contain the unexplored spaces as shown in Fig. \ref{frontier detection figure}d. The heuristic frontier regions $\mathcal{S}_{f}$ can be defined as a set containing all heuristic frontiers as follows: 
\begin{equation}
    \mathcal{S}_{f} = \{\mathcal{F}_{1}, \mathcal{F}_{2}, ..., \mathcal{F}_{n} \}, \ \mathcal{F}_{i} = \mathbb{C}_{i}(F_{i}, R_{i}, h_{i}),  
\end{equation}
where $\mathbb{C}$ represents a circle with its height $h$. In contrast to classic frontiers, our heuristic frontier regions serve as the direction guide for sampling and can be efficiently detected.

\textbf{Frontier Region Detection:} The process of detecting frontier regions is shown in Fig. \ref{frontier detection figure}. Given the robot's current 3D occupancy voxel map, we slice the voxel map into multiple 2D maps at different height levels (Fig. \ref{frontier detection figure}b). Then, based on the occupancy information from the 2D maps, we convert them into images (Fig. \ref{frontier detection figure}c), where grey pixels are the unexplored regions. Finally, we adopt the blob detection algorithm to the images to get the frontier regions. Note that we allow the heuristic frontier regions also to contain free and occupied spaces since they are primarily intended to guide the sampling process, as discussed in Section \ref{roadmap construction section}.

\subsection{Heuristic Incremental Roadmap Construction} \label{roadmap construction section}
\begin{figure*}[t]
    \centering
    \includegraphics[scale=0.525]{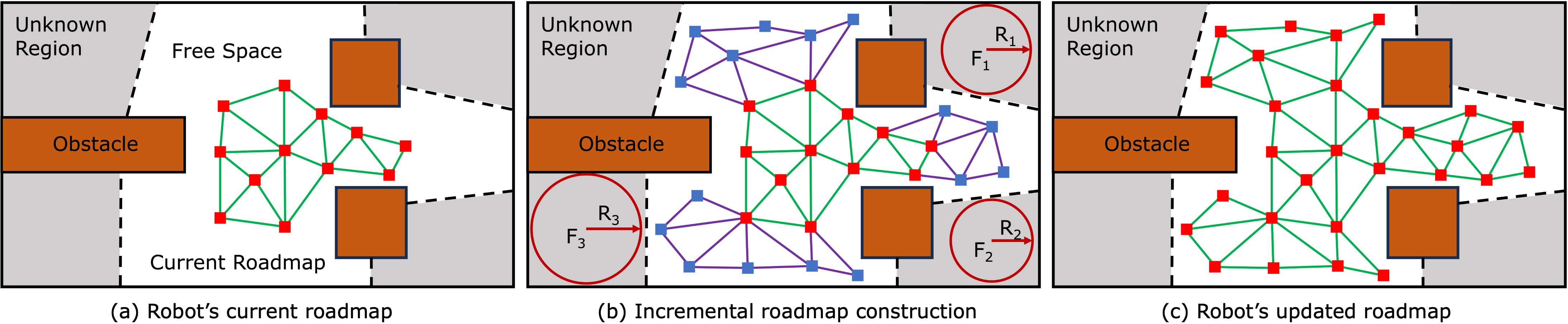}
    \caption{Heuristic incremental roadmap construction. With the robot's current roadmap shown in (a), the proposed algorithm first determines the heuristic frontier regions and then obtains sample nodes with edges toward the frontier regions depicted in (b). The updated roadmap is visualized in (c).}
    \label{incremental sampling figure}
\end{figure*}

\begin{algorithm}[t] \label{incremental roadmap algorithm}
\caption{Incremental Roadmap Construction} 
\SetAlgoNoLine%
$\mathcal{R} \gets \text{roadmap}$ \Comment*[r]{robot's current roadmap} \label{input start}
$\mathcal{S}_{f} \gets \{\mathcal{F}_{1}, \mathcal{F}_{2} ..., \mathcal{F}_{n} \}$ \Comment*[r]{frontier regions}
$\mathcal{M} \gets \text{occupancy map}$\;
$\xi_{r} \gets \text{robot current pose}$\; \label{input end}
$\mathcal{N}^{f}_{\text{fail}} \gets 0 $ \Comment*[r]{sampling failure number} \label{sample thresh start}
\While{$\mathcal{N}^{f}_{\normalfont{\text{fail}}} < \mathcal{N}_{\normalfont{\text{max}}}$}{  \label{sample thresh end}
    $n_{f} \gets \normalfont{\textbf{weightedSampleInFrontiers}}(\mathcal{S}_{f})$\; \label{frontier neighborhood start}
    $N \gets \mathcal{R}.\normalfont{\textbf{kNearestNeighbor}}(n_{f})$\; \label{frontier neighborhood end}
     \For{$n_{\text{i}}$ \normalfont{\textbf{in}} N}{
        $n_{i, ext} \gets \normalfont{\textbf{extendNodeToFrontier}}(n_{i}, n_{f})$\;
        \If{$\mathcal{M}.\normalfont{\textbf{isNodeFree}}(n_{i, ext})$}{ \label{start validity check}
            $nn_{i, ext} \gets \mathcal{R}.\normalfont{\textbf{nearestNeighbor}}(n_{i, ext})$\; \label{extend node}
            \If{$d_{\normalfont{\text{min}}} \leq  ||n_{i, ext} - nn_{i, ext}||_{2} \leq  d_{\normalfont{\text{max}}}$}{
                $\mathcal{R}.\normalfont{\textbf{insert}}(n_{i, ext})$ \Comment*[r]{add node}
            }
            \Else{
                $\mathcal{N}^{f}_{\text{fail}} \gets \mathcal{N}^{f}_{\text{fail}} + 1$ \; \label{end validity check}
            }
        }
    }
}
$\mathcal{R} \gets \normalfont{\textbf{localRegionSampling}}(\mathcal{R}, \mathcal{M}, \xi_{r})$\; \label{local sampling}
$\mathcal{R} \gets \normalfont{\textbf{globalRegionSampling}}(\mathcal{R}, \mathcal{M}, \xi_{r})$\; \label{global sampling}
$\mathcal{R}.\normalfont{\textbf{connectNode}}(\mathcal{M})$ \Comment*[r]{node connection} \label{node connection}
$\textbf{return} \ \mathcal{R}$\; 
\end{algorithm}

After detecting the heuristic frontier regions, we can use them to incrementally construct the probabilistic roadmap in each planning iteration. Our objective is to ensure the nodes in the roadmap can be evenly distributed within the free space and that the newly sampled nodes can make the roadmap grow toward the unexplored regions. The process of the incremental roadmap construction is shown in Fig. \ref{incremental sampling figure}, and the proposed incremental roadmap construction algorithm is presented in Alg. \ref{incremental roadmap algorithm}. Within each planning iteration, the algorithm will take the robot's current roadmap, frontier regions, occupancy map, and its current pose as inputs (Lines \ref{input start}-\ref{input end}). Note that we use the k-d tree data structure for the roadmap to achieve the efficient nearest neighbor search. Initially, we set a heuristic sampling failure number $\mathcal{N}^{f}_{\text{fail}}$ and draw samples until this value exceeds the threshold $\mathcal{N}_{\text{max}}$ (Lines \ref{sample thresh start}-\ref{sample thresh end}). For the roadmap node sampling, we begin by performing weighted sampling of the heuristic frontier $n_{f}$ in the heuristic frontier regions $\mathcal{S}_{f}$ and then find its neighborhood $N$ in the roadmap (Lines \ref{frontier neighborhood start}-\ref{frontier neighborhood end}). Next, for each neighbor $n_{i}$ in the neighborhood $N$, the candidate roadmap node $n_{i, ext}$ is obtained by extending the neighbor $n_{i}$ toward the heuristic frontier $n_{f}$ with a user-defined distance $\delta$ (Line \ref{extend node}). Finally, the validity check on the candidate node $n_{i, ext}$ will be applied to ensure the node is free and stays within the acceptable range to its nearest neighbor (Lines \ref{start validity check}-\ref{end validity check}). Notably, this validity check can ensure the roadmap nodes are evenly distributed. After this sampling stage, the roadmap can efficiently grow toward the unknown regions. 

Although the previously mentioned heuristic sampling stage can guide the roadmap's growing direction, we also want to build the roadmap in the rest of the free space for generating high-quality exploration paths. As a result, besides the heuristic sampling stage, the local and global sampling stages are also added (Lines \ref{local sampling}-\ref{global sampling}). For the local sampling stage, we randomly draw sample nodes in the predefined sample region and then perform the validity check as Lines \ref{start validity check}-\ref{end validity check}. If the sampled node passes the check, we add the node to the roadmap. Otherwise, we increment the failure count and exit the sampling stage when the failure count exceeds a threshold. For the global sampling stage, we follow a similar manner as the local counterpart but only change the sampling range. After getting all new samples, we perform the node connection operation (Line \ref{node connection}), which builds the collision-free edges to each node's neighborhood. 

\subsection{Node Pruning \& Information Update} \label{information section}
\begin{figure}[t]
    \centering
    \includegraphics[scale=0.375]{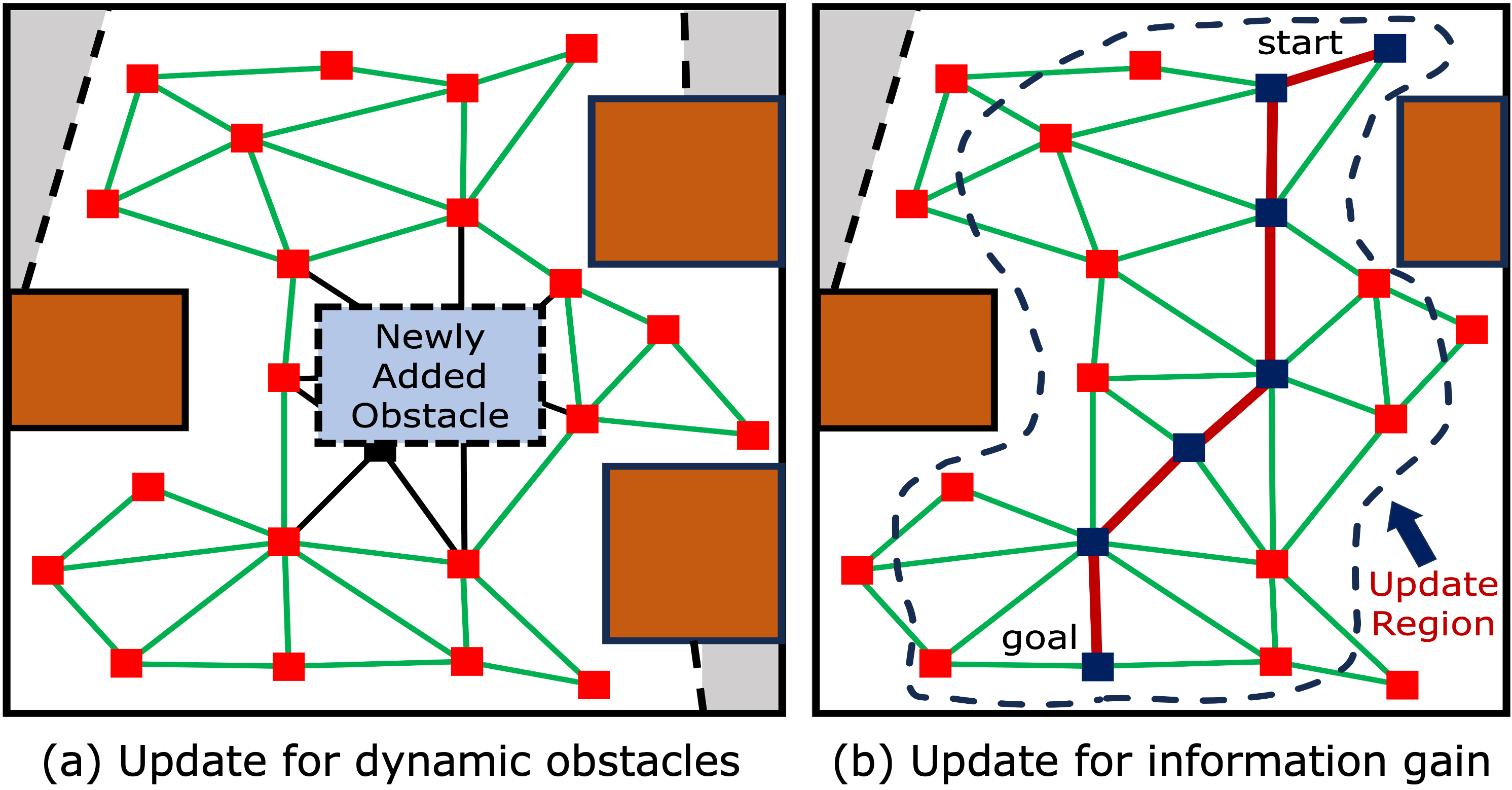}
    \caption{Illustration of the node information update. (a) Pruning invalid nodes and edges, represented in black, due to newly added obstacles. (b) Updating nodes' information gain after completing the exploration path. }
    \label{node update figure}
\end{figure}

Since the exploration environments can be dynamic, some previously valid nodes and edges from the incrementally built roadmap can become invalid due to the changing position of obstacles. For example, new obstacles can be added to the environment, as depicted in Fig. \ref{node update figure}a. As a result, after the incremental roadmap construction process discussed in the previous section, we will go through nodes and edges within the roadmap to check their occupancy information and prune them if they become invalid due to environmental changes. 

Different from the traditional probabilistic roadmap, in which each node only contains position and edge information, our roadmap's nodes also contain the information gain for the exploration guide. The information gain function for the $i$th node in the roadmap in the view direction angle $\phi$ is:
\begin{equation}
    \mathbf{IG}(n_{i}, \phi) = |\{v| v \in V_{\text{unk}}, v \in \normalfont{\textbf{sensorRange}}(n_{i}, \phi)\}|,
\end{equation}
where the sensor range function returns the onboard camera's range positioned at node $n_{i}$ facing at direction angle $\phi$, and the information gain counts the unknown voxel number in the sensor range for the node at the view angles. For reducing computation, we discrete the view angles to store the information gain and use interpolation to find the information gain at any angle. With the process of environment exploration, the node information gain needs to be updated. So, in each planning iteration, we collect newly sampled nodes and the nodes that are closed to the robot path to perform information gain evaluation and update as shown in Fig. \ref{node update figure}b. After information gain updating, the updated nodes will be inserted into a priority queue $\mathbb{PQ}_{\text{ig}}$ with the higher information gain (at the highest information gain angle) node on the top for path selection. With node pruning and updating, we can ensure the roadmap always has collision-free nodes with updated information gain data.

\subsection{Path Selection \& Trajectory Generation} \label{path section}
\begin{algorithm}[t] \label{path selection algorithm}
\caption{Path Selection and Traj. Generation} 
\SetAlgoNoLine%
$\mathcal{R} \gets \text{roadmap}$ \Comment*[r]{robot's current roadmap} 
$\mathbb{PQ}_{\text{ig}} \gets \text{information gain priority queue}$\; \label{priority queue}
$\mathbf{IG}_{\text{curr}} \gets 0, \  \mathbf{IG}_{\text{max}} \gets \normalfont{\textbf{highestInfoGain}}(\mathbb{PQ}_{ig})$\;
$S_{\sigma, \text{max}} \gets 0, \sigma^{\text{best}}_{\text{wp}} \gets \emptyset$\; 
\While{$\mathbf{IG}_{\normalfont{\text{curr}}} > \alpha \cdot \mathbf{IG}_{\normalfont{\text{max}}} \  \normalfont{\textbf{and} \  \mathbb{PQ}_{\text{ig}}}.\normalfont{\textbf{size}}() \neq 0 $}{ \label{while loop}
    $n_{\normalfont{\text{gc}}}, \mathbf{IG}_{\normalfont{\text{curr}}} \gets \mathbb{PQ}_{\text{ig}}.\normalfont{\textbf{top}}()$\;  \label{start find path}
    $\sigma_{\normalfont{\text{wp}}} \gets \normalfont{\textbf{shortestPathSearch}}(n_{\normalfont{\text{gc}}}, \mathcal{R})$\;
    $S_{\sigma} \gets \normalfont{\textbf{evaluatePathScore}}(\sigma)$\; \label{evaluate path score}
    \If{$S_{\sigma} > S_{\sigma, \normalfont{\text{max}}}$}{
        $S_{\sigma, \text{max}} \gets S_{\sigma}$\;
        $\sigma^{\text{best}}_{\text{wp}} \gets \sigma_{\normalfont{\text{wp}}}$ \Comment*[r]{best waypoint path} \label{end find path}
    }
    $\mathbb{PQ}_{\normalfont{\text{ig}}}.\normalfont{\textbf{pop}}()$\; 
}
$\sigma_{\text{traj}} \gets \normalfont{\textbf{optimizeTrajectory}}(\sigma^{\text{best}}_{\text{wp}})$\; \label{trajectory optimization}

$\normalfont{\textbf{return}} \  \sigma_{\text{traj}}$\;
\end{algorithm}
This section introduces the path selection and trajectory generation method (Alg. \ref{path selection algorithm}) based on our probabilistic roadmap. After constructing the roadmap with node information update, we can obtain the node order based on its information gain value from the priority queue $\mathbb{PQ}_{\text{ig}}$ (Line \ref{priority queue}), with the highest on the top. Then, we traverse through all nodes from the top node to the node value with information gain value less than $\alpha$ times the highest information gain $\textbf{IG}_{\text{max}}$, where $\alpha$ is a user-defined threshold (Line \ref{while loop}). For each node, we search the shortest path based on the roadmap, then evaluate the path score and find the best waypoint path $\sigma^{\text{best}}_{\text{wp}}$ with the highest score $S_{\sigma, \text{max}}$ (Lines \ref{start find path}-\ref{end find path}). The path score is the information gain rate of the path consisting of each waypoint node at the moving angle to its next waypoint:
\begin{equation}
    S_{\sigma} = \frac{1}{t} \sum^{N}_{i=0} \textbf{IG}(n_{i}, \textbf{atan2}(\frac{n_{i+1}.y - n_{i}.y}{n_{i+1}.x - n_{i}.x})), \ n_{i} \in \sigma_{\text{wp}},
\end{equation}
where $t$ is the estimated path execution time, and note that the last node's angle is its highest information gain angle. After getting the best exploration path, we apply the vision-aided B-spline trajectory \cite{visionaided} to navigate waypoints of the paths and avoid dynamic obstacles (Line \ref{trajectory optimization}). In some cases, the dynamic obstacles might occupy the path waypoints, resulting in the failure of the trajectory optimization. In this case, the proposed planner will re-select a path for exploration. Since the nodes and edges of the roadmap are guaranteed to be collision-free, our path search does not need to perform the collision checking operation, making the best path replanning extremely fast for collision avoidance. 

\section{Result and Discussion}
To assess the performance of the proposed planner, we conduct a series of experiments, including simulations and physical flight tests in various environments. The algorithm is implemented using C++ and ROS. For the simulation experiments, we utilize Gazebo/ROS and run the algorithm on Intel I7-12700k@3.8GHz. The physical flight tests are conducted using our customized quadcopter, equipped with an Intel RealSense D435i camera and an NVIDIA Orin NX onboard computer. We apply the visual-inertial odometry (VIO) algorithm \cite{vins} for robot state estimation and the dynamic obstacle detector \cite{dynamic_map} for collision avoidance. The parameters and settings for the robot in the simulation experiments are detailed in Table \ref{setting table}. For the physical flight tests, we lower the maximum linear velocity limits to 0.5m/s and the maximum angular velocity to 0.5rad/s, respectively.

\begin{table}[h]
\begin{center}
\caption{Robot parameters and settings.} \label{experiment_parameters}

\begin{tabular}{l  l  l  l} 
\hline

Max. Linear Vel. & 1.0 m/s & Collision Box & [0.5, 0.5, 0.3]$\text{m}^\text{3}$\Tstrut\\

Max Angular Vel.  & 0.8 rad/s & Camera Range & 5 m\Tstrut\\ 

Map Resolution & 0.1 m & Camera FOV & [86, 57]\degree \Tstrut\\  

\hline
\end{tabular}\label{setting table}
\end{center}

\end{table}

\begin{figure}[t]
    \centering
    \includegraphics[scale=0.74]{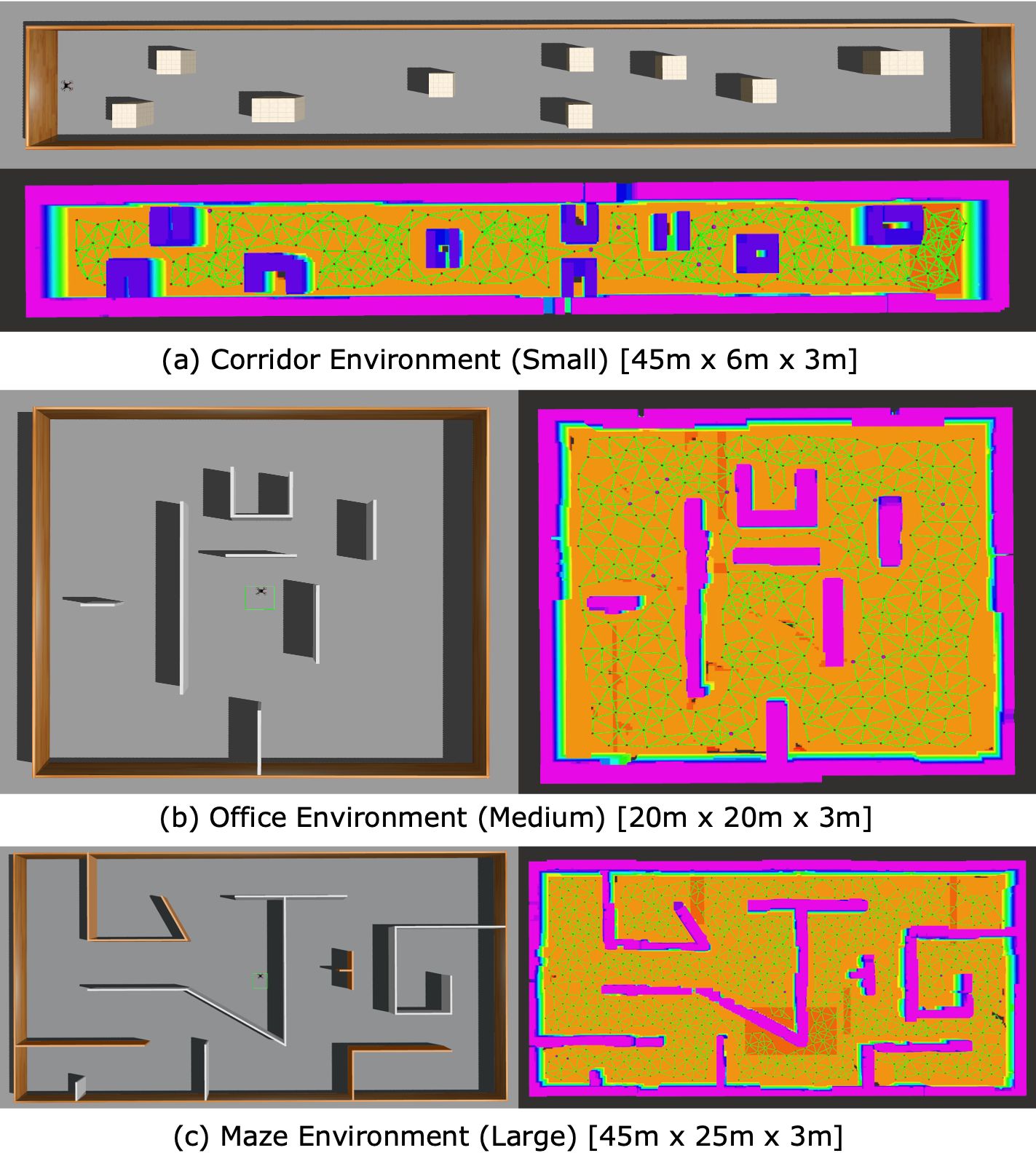}
    \caption{Visualization of the simulation environments (Corridor, Office, and Maze) with their fully explored map built by the proposed algorithm.}
    \label{simulation map figure}
\end{figure}

\subsection{Simulation Experiments:}
\begin{figure*}[t]
    \centering
    \includegraphics[scale=0.465]{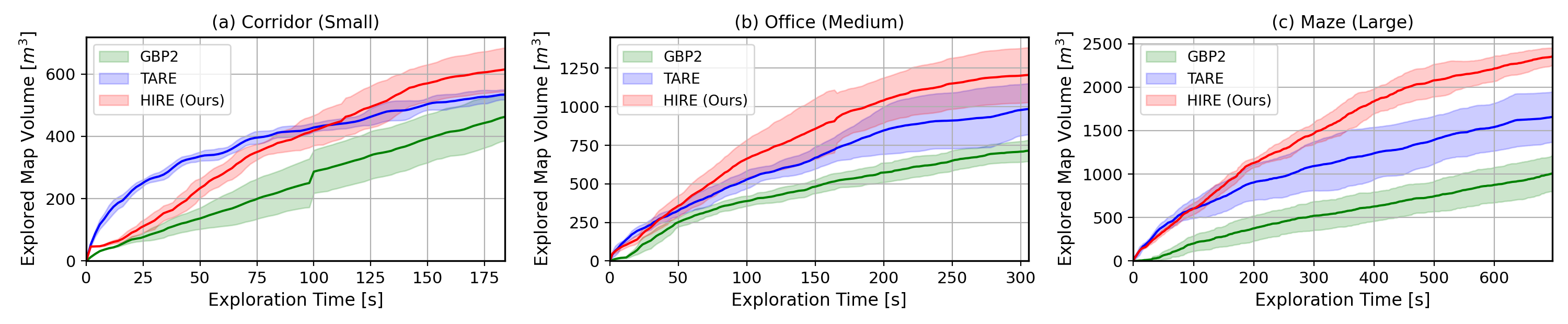}
    \caption{Comparison of the average exploration rates in simulation environments: Corridor (Small), Office (Medium), and Maze (Large). The shaded areas show the one-standard-deviation ranges of the exploration rates. Our planner achieves the highest exploration rate across three environments.}
    \label{exploration rate figure}
\end{figure*}

To evaluate the exploration efficiency, we perform the benchmarking experiments against two state-of-the-art exploration planners \cite{GBP2}\cite{TARE} in simulation environments. We prepare three simulation environments of different scales: corridor (small), office (medium), and maze (large), as shown in Fig. \ref{simulation map figure}. From the figure, one can see that our planner is able to fully explore environments of different scales. To compare the exploration performance, we recorded the exploration rate, the trajectory length, and the exploration time of three planners by running each algorithm 10 times in each environment. The comparison of the exploration rates with one-standard-deviation ranges between planners is visualized in Fig. \ref{exploration rate figure}.
It's evident that our planner consistently maintains the highest exploration rate throughout most of the exploration time across all three environments. Moreover, as the size of the environments increases, our planner's performance surpasses that of the other two planners even more noticeably. Based on our experimental observations, this performance improvement can be attributed to the extensive coverage of the probabilistic roadmap, particularly in the frontier areas due to the heuristic sampling. This comprehensive coverage results in more efficient exploration trajectories. Among the benchmarking algorithms, we noticed a higher occurrence of back-and-forth trajectories during exploration in the larger-sized environments. In addition to the exploration rate, we provide a comparison of trajectory length and exploration time for the three planners required to fully explore the three environments in Table \ref{length and time comparison table}. As demonstrated in the table, our proposed planner exhibits the shortest trajectory length and the fastest exploration time across all three environments.

\begin{table}[h]
\centering
\caption{Each planner's average trajectory length and the exploration time with standard deviations in simulation.}
\begin{tabular}{|c|c|c|c|c|c|}
\hline
Env. Name & Planner & Traj. Length (m) & Expl. Time (s) \Tstrut\\
\hline
 & GBP2 \cite{GBP2} & $265.2 \pm 20.2$ & $301.2 \pm 14.9$ \Tstrut\\
Corridor        & TARE\cite{TARE}            & $233.8 \pm 17.8$ & $289.6 \pm 23.7$  \Tstrut\\                
                 & \textbf{HIRE} & $\textbf{192.8} \pm \textbf{16.5}$ & $\textbf{244.8} \pm \textbf{28.5}$ \Tstrut\\
\hline
 & GBP2 \cite{GBP2} & $366.7 \pm 30.6$ & $486.3 \pm 34.6$ \Tstrut\\
Office        & TARE \cite{TARE}           & $312.3 \pm 29.9$ & $406.1 \pm 38.6$ \Tstrut\\                
                 & \textbf{HIRE} &$\textbf{231.9} \pm \textbf{24.9}$ & $\textbf{310.8} \pm \textbf{32.6}$ \Tstrut\\
\hline
 & GBP2 \cite{GBP2} & $1132.2 \pm 50.5$ & $1426.8 \pm 70.3$ \Tstrut\\
Maze       & TARE \cite{TARE}            & $701.3 \pm 35.5$ & $912.8 \pm 52.3$ \Tstrut\\                
                 & \textbf{HIRE} & $\textbf{427.4} \pm \textbf{40.2}$ & $\textbf{712.3} \pm \textbf{46.3}$ \Tstrut\\
\hline
\end{tabular}
\label{length and time comparison table}
\end{table}

\begin{figure*}[t]
    \centering
    \includegraphics[scale=0.535]{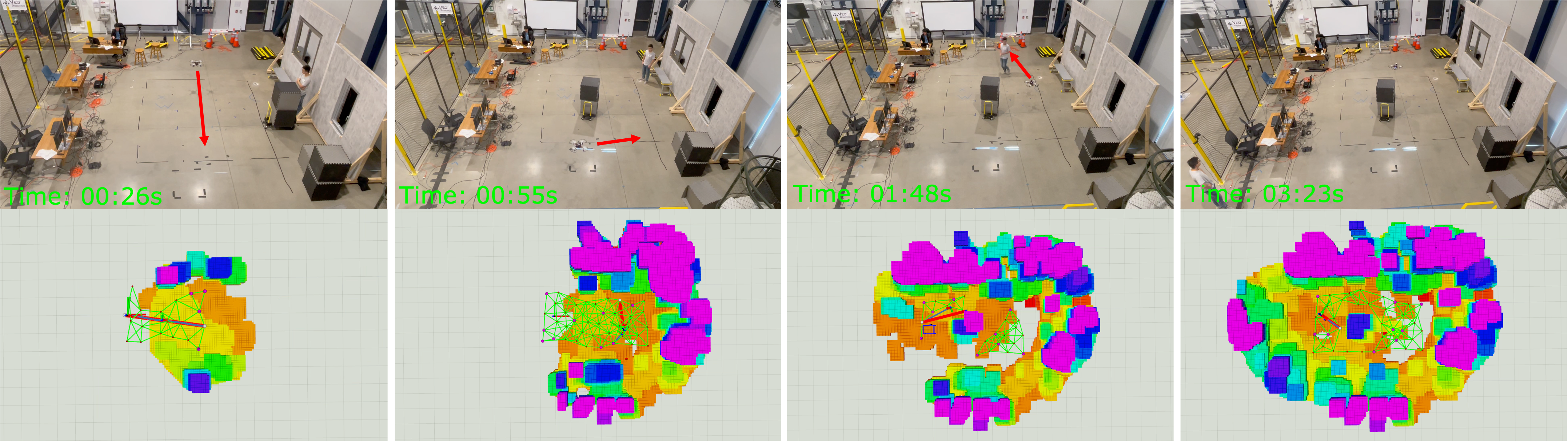}
    \caption{Illustration of a physical exploration flight experiment in a dynamic environment. The top and bottom figures display the environment and the Rviz visualization of the current mapped area using the incremental probabilistic roadmap, respectively. The environment includes walking pedestrians, and we also modify static obstacle positions while the robot is exploring. The robot can fully explore the environment while safely avoiding obstacles.}
    \label{flight experiment figure}
\end{figure*}

\begin{table}[h]
    \centering
    \caption{The runtime of each component of the proposed planner measured using the NVIDIA Orin NX onboard computer.}
    \begin{tabular}{ c c c } 
    \hline
    Planner Components & Run Time (ms) & Portion (\%) \Tstrut\\
    \hline
    Heuristic Frontier Detection & 3.25 & 4.22\% \Tstrut\\ 
    Incremental Roadmap Construction & 25.50 & 33.12\% \\ 
    Node Pruning \& Info. Update & 10.50 & 13.63\% \\ 
    
    Path Selection \& Traj. Generation & 37.75 & 49.03\% \\ 
    
    \textbf{Planner Total Time} & \textbf{77.00} & \textbf{100.00}\% \\
    \hline

    \end{tabular}
    \label{run time table}
\end{table}

\subsection{Physical Flight Test:}
To test the robustness of our planner, we perform physical flight tests in different dynamic environments, as shown in Fig. \ref{intro figure} and Fig. \ref{flight experiment figure}. One of the test field measures 10m in length, 6m in width, and 3m in height and an illustration of one such physical flight test is presented in Fig. \ref{flight experiment figure}. In the figure, it is evident that the explored area, as depicted in Fig. \ref{flight experiment figure} bottom, steadily expands over time, allowing the robot to thoroughly explore the entire environment within 3.5 minutes. During our experiments, we introduced two categories of dynamic obstacles into the environments. The first type involved continuously moving dynamic obstacles, where a human pedestrian walks through the environment, prompting the robot to detect and dynamically replan a collision-free trajectory for collision avoidance, as illustrated in Fig. \ref{flight experiment figure}c. The second type of dynamic obstacles entailed alterations to the static environment by adding or removing static obstacles. This adjustment triggered updates and the initiation of node pruning within the explored region. As depicted in Fig. \ref{flight experiment figure}a-b and Fig. \ref{flight experiment figure}d, we introduced a new obstacle into the center of the environment. The experiment demonstrates the successful execution of both replanning and node pruning in real-world scenarios with dynamic obstacles. The final exploration trajectory length is 32.33m. The average recorded runtime of the proposed planner for the robot onboard computer is presented in Table \ref{run time table}. The total planner runtime is 77ms, which can run over 10Hz. Notably, the heuristic frontier detection only takes 3.25ms, which is 4.22\% of the total planner runtime. The remaining components all operate within a computational time of less than 50ms, ensuring the planner's real-time performance.

\section{Conclusion and Future Work}
This paper introduces the Heuristic-based Incremental Probabilistic Roadmap Exploration (HIRE) planner designed for UAV exploration in dynamic environments. The proposed algorithm first adopts a computationally efficient vision-based algorithm to detect the heuristic frontier regions from different map layers. Then, it applies the incremental sampling strategy to construct the roadmap for exploration. Additionally, our dynamic module ensures timely updates to roadmap information to accommodate dynamic environments, facilitating the efficiency of exploration and safety of dynamic obstacle avoidance. The simulation experiments show that the proposed method outperforms the state-of-the-art benchmark algorithms' exploration efficiency. Furthermore, our physical flight test demonstrates that our algorithm can make UAVs explore safely in dynamic environments. Our future work will focus on the heterogeneous robot team exploration with the incremental probabilistic roadmap.

\bibliographystyle{IEEEtran}
\bibliography{bibliography.bib}

\end{document}